\documentclass[10pt,twocolumn,letterpaper]{article}

\usepackage{iccv}              % To produce the CAMERA-READY version
\usepackage{amssymb}
\usepackage{xcolor}
\usepackage{pifont}
\definecolor{iccvblue}{rgb}{0.21,0.49,0.74}
\usepackage[pagebackref,breaklinks,colorlinks,allcolors=iccvblue]{hyperref}

\def\paperID{14247} % *** Enter the Paper ID here
\def\confName{ICCV}
\def\confYear{2025}

\title{From Easy to Hard: The MIR Benchmark for Progressive Interleaved Multi-Image Reasoning}

\author{
    Hang Du\textsuperscript{\rm 1}\footnotemark[1],
    Jiayang Zhang\textsuperscript{\rm 1}\footnotemark[1],
    Guoshun Nan\textsuperscript{\rm 1}\footnotemark[2],
    Wendi Deng\textsuperscript{\rm 1},
    Zhenyan Chen\textsuperscript{\rm 1},\\
    Chenyang Zhang\textsuperscript{\rm 1},
    Wang Xiao\textsuperscript{\rm 1},
    Shan Huang\textsuperscript{\rm 1},
    Yuqi Pan\textsuperscript{\rm 1},
    Tao Qi\textsuperscript{\rm 1},
    Sicong Leng\textsuperscript{\rm 2}
    \and  
  \textsuperscript{\rm 1}Beijing University of Posts and Telecommunications
  \textsuperscript{\rm 2}Nanyang Technological University \and   
 \tt\small \{7597892, zzzjy, nanguo2021, dengwendi02, zhenyanchen\}@bupt.edu.cn, \and
 \tt\small\{2650465680, xiaowangsama, huangshan2022, panyuqi2022,\}@bupt.edu.cn, \and 
 \tt\small\{taoqi.qt, Lengsicong\}@gmail.com\\
    }
\begin{document}
\maketitle
\renewcommand{\thefootnote}{\fnsymbol{footnote}} 
\footnotetext[1]{Equal Contribution}
\footnotetext[2]{Corresponding Author}
\begin{abstract}
\label{Sec:abs}
% Multi-image Interleaved reasoning (MIR) aims to enable MLLMs (Multimodal Large Language Models) to comprehend and reason across multiple images and their accompanying textual information, addressing the limitations of single-image or non-interleaved multi-image tasks. 
Multi-image Interleaved Reasoning aims to improve Multimodal Large Language Models' (MLLMs) ability to jointly comprehend and reason across multiple images and their associated textual contexts, introducing unique challenges beyond single-image or non-interleaved multi-image tasks.
% Current multi-image benchmarks lack interleaved textual data, limiting their ability to fully explore the intricate relationships between text and images.
% Our work demonstrates the significant impact of multi-image interleaved data in enhancing models' ability to understand complex scenes, capture cross-modal correlations, and improve reasoning capabilities.
While current multi-image benchmarks overlook interleaved textual contexts and neglect distinct relationships between individual images and their associated texts, enabling models to reason over multi-image interleaved data may significantly enhance their comprehension of complex scenes and better capture cross-modal correlations.
% To bridge this gap, we propose a novel benchmark for Multi-Image Reasoning (MIR), which raises critical questions: ``How to accurately match specific image regions with their corresponding texts?'' and ``How to establish logical associations between multiple-image regions?''.
% Specifically, our benchmark focuses on multi-image interleaved data, each instance requiring reasoning across multiple images accompanied by interleaved texts.
To bridge this gap, we introduce a novel benchmark \textbf{MIR}, requiring joint reasoning over multiple images accompanied by interleaved textual contexts to accurately associate image regions with corresponding texts and logically connect information across images.
% Our method guides the model through a progressive learning path, starting with simple samples and advancing to complex ones, while gradually increasing the complexity of the tasks. 
% The "easy to hard" approach enables the model to progressively master challenging tasks, enhancing its reasoning capabilities and generalization performance.
% Furthermore, to address these challenges, we construct instruction-tuning data with structured reasoning annotations for each instance and propose a stage-wise curriculum learning strategy.
To enhance MLLMs' ability to comprehend multi-image interleaved data, we introduce reasoning steps for each instance within the benchmark and propose a stage-wise curriculum learning strategy. 
This strategy follows an ``easy to hard'' approach, progressively guiding models from simple to complex scenarios, thereby enhancing their ability to handle challenging tasks.
Extensive experiments benchmarking multiple MLLMs demonstrate that our method significantly enhances models' reasoning performance on MIR and other established benchmarks.
% highlighting the challenges current MLLMs face with multi-image interleaved reasoning
% We believe that MIR will inspire the community to enhance MLLMs, bridging the gap in handling interleaved image-text data.
We believe that MIR will encourage further research into multi-image interleaved reasoning, facilitating advancements in MLLMs' capability to handle complex inter-modal tasks.
Our code and dataset are available at \url{https://github.com/Shelly-coder239/MIRBench}.
\end{abstract}

% The ABSTRACT is to be in fully justified italicized text, at the top of the left-hand column, below the author and affiliation information.
% Use the word ``Abstract'' as the title, in 12-point Times, boldface type, centered relative to the column, initially capitalized.
% The abstract is to be in 10-point, single-spaced type.
% Leave two blank lines after the Abstract, then begin the main text.
% Look at previous \confName abstracts to get a feel for style and length.

% because reasoning requires not only understanding the input but also maintaining consistency and logical clarity across multiple steps.    
\section{Introduction}
\label{Sec:intro}
\begin{figure}[ht!]
    \centering
\includegraphics[width=1.0\linewidth]{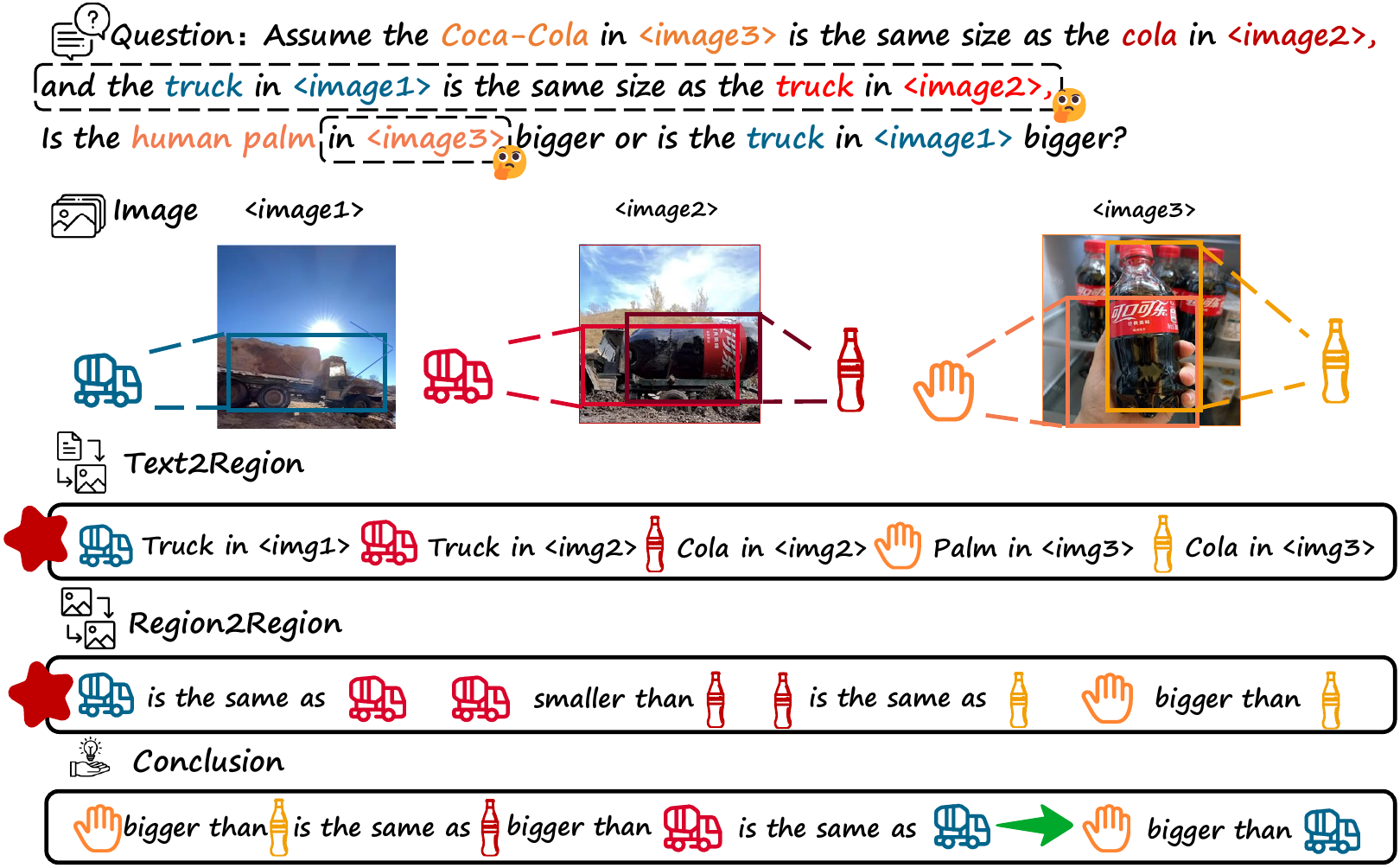}
    \caption{\textbf{An example from MIR.} This question aims to compare the sizes of the ``truck'' and ``palm''. Dashed box content is inferred by the model from context. \textit{Text2Region} connects explicit textual elements (``truck'' and ``cola'') and  implicit objects (``human palm in image3'', ``truck in image1'') to image regions. \textit{Region2Region} establishes relationships between these regions, such as the cola and truck in image2, the palm and cola in image3. By reasoning over the relationship between ``palm'' ``Coca-Cola'', and ``truck'', the correct size relationship is derived.}
    \vspace{-3mm}
    \label{fig:introexample}
\end{figure}
Interleaved image-text data refers to a sequence in which multiple images and text segments are dynamically interleaved, creating a cohesive flow that integrates both visual and textual information \cite{tian2024mm, InterleavedScene, CoMM}.
Textual content alone often struggles to fully convey complex or nuanced messages\cite{MC4, OBELISC, Kosmos, MIMICIT, DataComp}. 
By interleaving multiple images with text, the integration of visual elements significantly enriches information delivery\cite{Impact, YUM2021101397, 9578044, cuva}. 
% This approach not only improves accessibility and comprehensibility but also allows for a more dynamic and contextually rich representation of information. \cite{}.
\begin{figure*}[ht!]
    \centering
    \includegraphics[width=180mm]{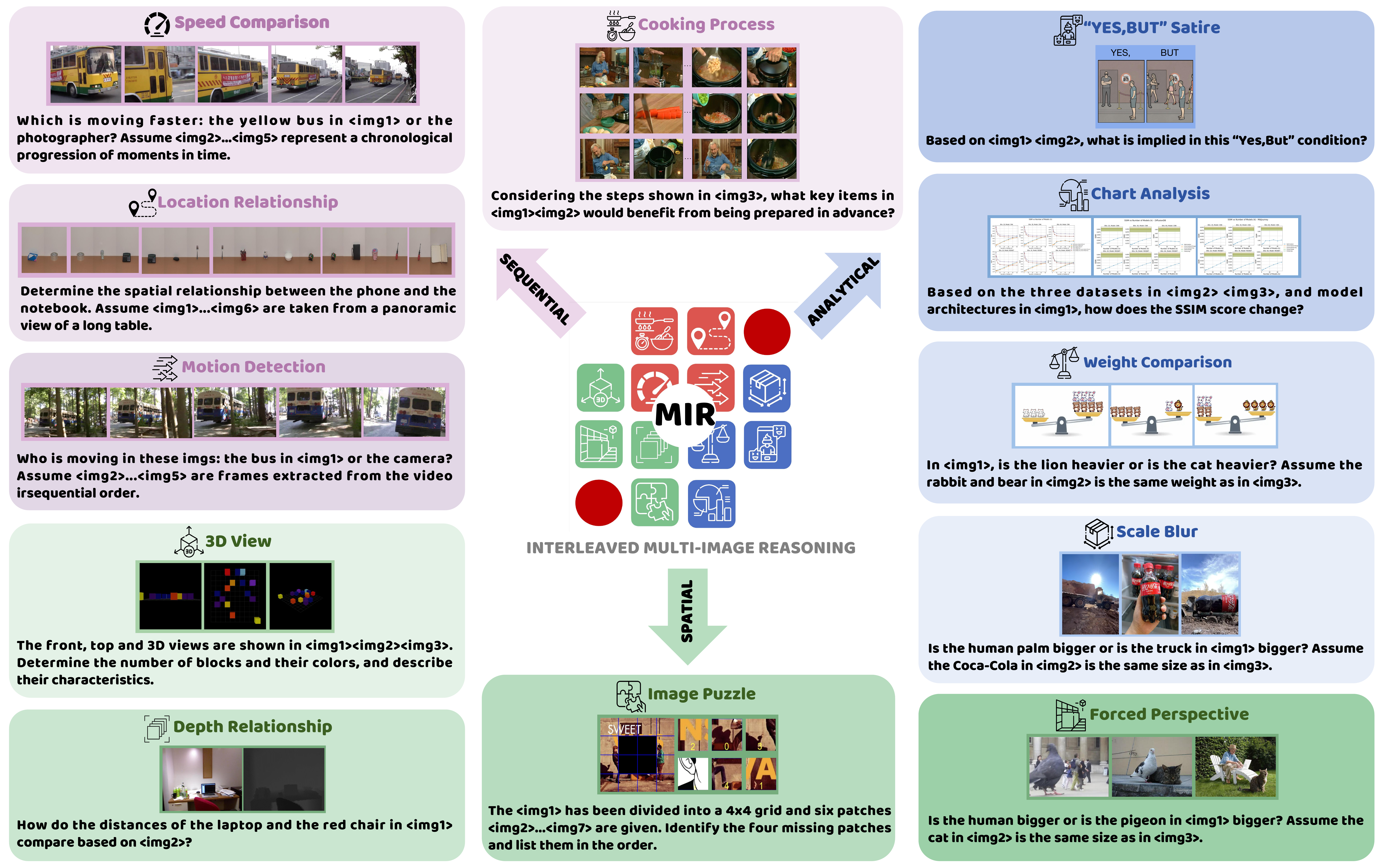}
    \caption{\textbf{Overview of our proposed MIR Benchmark.} It comprises $22,257$ questions derived from $138,277$ images, organized into three distinct categories—sequential ($5,455$), spatial ($9,350$), and analytical ($7,821$)—which are further divided into $12$ fine-grained tasks to rigorously evaluate multi-image interleaved reasoning.}
    \label{fig:intromain}
    \vspace{-5mm}
\end{figure*}

Multi-image interleaved data, widely prevalent across social media platforms, news outlets, and digital publications\cite{ALWASHMI2024120483, Visualaids, REN2025156, ROLEOFVISUAL}, plays a crucial role in a wide range of applications within journalism and related fields.\cite{docmsu, yesbut,ISHMAM2024102270, SHARMA2022116159,nan-etal-2020-reasoning, 10963886}.

\indent Recent advancements in multimodal large language models (MLLMs) have demonstrated remarkable capabilities across diverse multimodal contexts\cite{whoops}.
However, previous work primarily focused on pushing the performance limits of single-image or multi-image tasks\cite{meng2024mmiu,yan2025mir,fu2024blink,wang2024muirbench}, with limited exploration of multi-image interleaved scenarios.
Multi-image interleaving tasks require MLLMs to exhibit precise image comprehension, accurately align images with corresponding texts, and establish correct associations between them.
Figure \ref{fig:introexample} illustrates a multi-image interleaved Q\&A example.
Accurately answering the question about the size ratio between the truck in \textit{image1} and the palm in \textit{image3} involves two key challenges: \textit{Text2Region} and \textit{Region2Region}.
First, \textit{Text2Region} requires matching explicit textual elements like ``truck'' and ``cola'' to their corresponding images (\textit{image1}, \textit{image2} and \textit{image3}). 
It also needs to implicitly link contextually inferred objects, such as `` truck'' and ``palm'' to their respective images (\textit{image1} and \textit{image3}).
Next, \textit{Region2Region} involves establishing the relationships between these regions, such as comparing the ``truck'' and the ``cola'' in \textit{image2}, the ``cola'' and the ``palm'' in \textit{image3}. 
By establishing the logical chain connecting ``palm'', ``Coca-Cola'', and ``truck'', the correct conclusion can be derived. 
These challenges necessitate the development of MLLMs that explicitly account for the unique characteristics of interleaved multi-image data.

\indent Previous studies have highlighted the critical role of large, high-quality, and challenging benchmarks in developing and evaluating MLLMs for multi-image reasoning tasks.
Building on this foundation, existing benchmarks have demonstrated their potential \cite{BLINK, llava-next-interleave, wang2024muirbench}.
However, they still exhibit limitations in addressing more practical real-world scenarios:
1) \textit{Limited evaluation of multi-image reasoning.} Current multi-image benchmarks primarily focus on basic multi-image question-answering, often requiring only the extraction of textual information from images and relying on LLMs for text-based reasoning. Consequently, they fail to incorporate interleaved textual information and overlook more complex reasoning paradigms, such as text-to-region and region-to-region inference, which are essential for capturing the deeper interplay between visual and textual content.\cite{meng2024mmiu, wu2024mmra, song2024milebench, Img-diff, li2024naturalbench}.
2) \textit{Lack of step-by-step reasoning.} Existing multi-image benchmarks focus primarily on end-task accuracy, neglecting the intermediate reasoning steps.
3) \textit{Data leakage.} Since most of the existing datasets are derived from other widely used public datasets, which have already been pre-trained by MLLMs, this could potentially result in data leakage issues\cite{llava-next-interleave,jiang2024mantis}.\\
\indent These limitations highlight the need for a benchmark that not only integrates interleaved multi-image data but also supports advanced reasoning capabilities.
To this end, we introduce MIR, a comprehensive benchmark consisting of 22,257 challenging image-text interleaved QA pairs across 12 diverse scenarios.
Our dataset includes $138,277$ images, averaging six images per instance, as illustrated in Figure \ref{fig:intromain}.
We equip each instance with $5$ systematic and structured reasoning steps:
\textit{Summary}, \textit{Caption}, \textit{Text to region reference}, \textit{Region to Region relationship}, \textit{Conclusion}.
Based on reasoning steps, we propose a stage-wise curriculum learning approach, which prioritizes training on simple samples and gradually introduces harder ones. 
When training on harder samples, we leverage reasoning steps to continuously guide the MLLMs in mastering the reasoning process.
Through step-wise training schema, our method guides the MLLMs to progressively master complex tasks from ``easy'' to ``hard'', thereby enhancing the model's reasoning capabilities.

\begin{table*}[htb!]
    \centering
    \resizebox{1.0\linewidth}{!}{
    \begin{tabular}{l|r|r|c|c|c|c|c|c}
    \hline
        \textbf{Dataset} & \textbf{Samples} & \textbf{Images} & \textbf{Interleave} & \textbf{Multi-image} & \textbf{Rationale} & \textbf{Steps} & \textbf{Answer-type} & \textbf{Annotator} \\ \hline
    MVbench\cite{li2024mvbench} & 4,000 & 4,000 & \textcolor{blue}{\checkmark} & \textcolor{red}{\ding{55}} & \textcolor{red}{\ding{55}} & \textcolor{red}{\ding{55}} & MCQ & Auto \\ \hline
        MMIU~\cite{meng2024mmiu} & 11,600 & 77,000 & \textcolor{red}{\ding{55}} & \textcolor{blue}{\checkmark} & \textcolor{blue}{\checkmark} & \textcolor{red}{\ding{55}} & MCQ & Auto \\ \hline
        MIRBENCH~\cite{yan2025mir} & 925 & 3,496 & \textcolor{red}{\ding{55}} & \textcolor{blue}{\checkmark} & \textcolor{red}{\ding{55}} & \textcolor{red}{\ding{55}} & MCQ \& Free-form & Auto \\ \hline
        BLINK~\cite{fu2024blink} & 3,807 & 7,300 & \textcolor{red}{\ding{55}} & \textcolor{blue}{\checkmark} & \textcolor{blue}{\checkmark} & \textcolor{red}{\ding{55}} & MCQ & Semi-Manual \\ \hline
        MUIRBench~\cite{wang2024muirbench} & 2,600 & 11,264 & \textcolor{blue}{\checkmark} & \textcolor{blue}{\checkmark} & \textcolor{blue}{\checkmark} & \textcolor{red}{\ding{55}} & MCQ & Semi-Manual \\ \hline
        LLAVA-INTERLEAVE~\cite{li2024llavanextinterleave} & 17,000 & - & \textcolor{blue}{\checkmark} & \textcolor{blue}{\checkmark} & \textcolor{red}{\ding{55}} & \textcolor{red}{\ding{55}} & MCQ \& Free-form & Auto \\ \hline
        MMRA~\cite{wu2024mmra} & 1,024 & 3,403 & \textcolor{red}{\ding{55}} & \textcolor{blue}{\checkmark} & \textcolor{red}{\ding{55}} & \textcolor{red}{\ding{55}} & MCQ & Semi-Manual \\ \hline
        MILEBench~\cite{song2024milebench} & 6,440 & 97,855 & \textcolor{red}{\ding{55}} & \textcolor{blue}{\checkmark} &  \textcolor{red}{\ding{55}} & \textcolor{red}{\ding{55}} & MCQ \& Free-form & Auto \\ \hline
        MMMU~\cite{yue2024mmmu} & 11,500 & 12,507 & \textcolor{blue}{\checkmark} & \textcolor{blue}{\checkmark} & \textcolor{blue}{\checkmark} & \textcolor{red}{\ding{55}} & MCQ \& Free-form & Semi-Manual \\ \hline
        MMIE\cite{MMIE} & 20,103 & 26,535 & \textcolor{blue}{\checkmark} & \textcolor{blue}{\checkmark} & \textcolor{blue}{\checkmark} & \textcolor{red}{\ding{55}} & MCQ \& Free-form & Semi-Manual \\ \hline
        MIR(Ours) & 22,257 & 138,277 & \textcolor{blue}{\checkmark} & \textcolor{blue}{\checkmark} & \textcolor{blue}{\checkmark} & \textcolor{blue}{\checkmark} & MCQ & Semi-Manual \\ \hline
    \end{tabular}
    }
    \caption{\textbf{Comparison of Multi-image benchmarks.} Compared to the existing benchmarks, MIR possesses the largest number of data samples and the highest number of images. Each sample in MIR is accompanied by five reasoning steps. Additionally, MIR employs a multiple-choice question (MCQ) format to ensure standardized and consistent evaluation. }
    \label{tab:dataset_comparation}
\end{table*}

% MIRBENCH Auto?   MVbench Images?
Our contributions are summarized as follows:
\begin{itemize}
	\item We curate MIR, a novel benchmark focused on interleaved multi-image reasoning and featuring a structured five-stage reasoning process to enhance models' ability to reason across multiple images and textual contexts
	\item We propose a stage-wise curriculum learning method to improve MLLMs reasoning capabilities and generalization performance. This method helps the model progressively grasp complex reasoning processes by advancing from simple to difficult tasks. 
% to localize sarcasm objects in both texts and images based on our dataset, and introduce three evaluation matrix to accurately estimate the model's performance.
	\item We conduct extensive experiments on MIR. Results show that the MIR enables us to push the boundaries of current MLLMs' reasoning capabilities.
\end{itemize}

\section{Related Work}
\label{Sec:related_work}

\subsection{Interleaved Multi-image Benchmarks}
Recent advancements in MLLMs lead to the development of numerous image benchmarks, which are crucial for evaluating multimodal large language models (MLLMs). 
Existing benchmarks for MLLMs primarily fall into two categories: multi-image benchmarks \cite{li2024naturalbench,jiang2024mantis,meng2024mmiu,li2024finetuning} and interleaved benchmarks \cite{fu2024blink,li2024finetuning,suhr2019corpusreasoningnaturallanguage,jiang2024manyshotincontextlearningmultimodal}.
The multi-image benchmarks comprehensively evaluate the performance of MLLMs across multiple dimensions of multi-image tasks.
NaturalBench~\cite{li2024naturalbench} incorporates a comprehensive set of adversarial samples to produce ``blind'' responses that fail to properly utilize visual information.
Img-Diff\cite{Img-diff} is designed to enhance fine-grained image recognition of MLLMs.
Mantis-Eval~\cite{jiang2024mantis} and MMIU~\cite{meng2024mmiu} challenge MLLMs with more comprehensive multi-image data.
Nevertheless, multi-image benchmarks' lack of interleaved textual data limits their capacity to capture vision-language connections and comprehend multimodal contextual information.
The interleaved benchmarks primarily concentrate on image-text interleaved data.
BLINK~\cite{fu2024blink} focuses on core visual perception abilities with visual prompting. 
The ICL benchmarks~\cite{jiang2024manyshotincontextlearningmultimodal,shukor2024taskperformanceevaluatingreducing} provide a thorough assessment of their interleaved capabilities across few-shot to many-shot settings.
LLaVa-Next-Interleave~\cite{li2024llavanextinterleave, LLaVA-OneVision} adopts the interleaved multi-image data format to strengthen the cross-modal capability in LMMs. 
However, they still face challenges in step-by-step reasoning and potential data leakage. 
More detailed comparisons are available in Table \ref{tab:dataset_comparation}.

\subsection{Multi-modal Large Language Models}
% Traditional vision-language models capture and represent the visual reasoning process using neural-symbolic methods~\cite{choi2025towards,pmlr-v119-amizadeh20a}. With the development of LLMs, vision-language models (VLMs) utilize the reasoning capabilities of LLMs to solve multimodal tasks, such as visual question answering, cross-modal retrieval, and image generation. For stronger multi-image LLMs, mPLUG-Owl3~\cite{ye2024mplugowl3} integrates vision and language into a semantic space, and Mantis~\cite{jiang2024mantis} performs instruction tuning with academic-level resources. For more efficient images processing, VILA-1.5~\cite{liu2024nvila} adopts a “scale-then-compress” approach, and InterVL 2.5~\cite{chen2025expanding} leverages a Chain-of-Thought (CoT). For easier transfer learning across different scenarios, LLaVa-OneVision~\cite{li2024llavaonevision} consolidates models and vision representations into LLaVa-NeXT~\cite{li2024llavanextinterleave}. Nevertheless, the prior VLMs still exhibit shortcomings in reasoning, including the relationships among entities and 
% in-context multimodal training schema. For this reason, we propose XXX for stronger fine-grained multi-image reasoning.

The development of reasoning capabilities in Large Language Models (LLMs) has been a focal point of research\cite{LLM-re, Tot}. 
Early studies often relied on neural-symbolic methods, employing formal languages rather than natural language to explicitly model the reasoning process\cite{chiang-chen-2019-semantically, amini-etal-2019-mathqa, choi2025towards,pmlr-v119-amizadeh20a}. 
However, with the emergence of powerful LLMs, new approaches have been proposed by leveraging their inherent reasoning abilities\cite{6313337, Robust-reasoning}. 
Techniques such as Chain-of-Thought (CoT) prompting, which decompose complex problems into intermediate reasoning steps, have demonstrated potential in guiding LLMs to generate structured solutions\cite{yang2024buffer, Got}. 
These methods have been further extended to MLLMs\cite{jiang2025mme, zhang2023multicot}. 
Visual programming \cite{10204174} provides a modular neuro-symbolic system that utilizes computer vision models as functions and integrates CoT techniques with GPT-3 LLM for compositional visual reasoning.
However, CoT-guided multi-step reasoning still struggles with consistency, often leading to errors and hallucinations\cite{thawakar2025llamavo1}. 
Some methods \cite{2024arXiv241016198Z,chen-etal-2024-measuring} leverage GPT-4 to generate reasoning process and incorporate both correct and incorrect reasoning chains during training, enhancing the model's reasoning capabilities through reinforcement learning (RL). 
Nevertheless, RL-based reasoning is slow and resource-intensive\cite{2024arXiv240400282C}. 
In contrast, our approach employs a stage-wise reasoning strategy, guiding the model from ``easy'' to ``hard'' tasks, which proves to be a more efficient approach for improving reasoning capabilities.
\section{Dataset}
\label{Sec:Dataset}
In this section, we first give a brief overview of MIR in Section \ref{dataset-overview}.
Then we describe each task in detail, providing an in-depth explanation of the data curation process in Section \ref{dataset-construction}.
The quantitative analysis of the MIR is illustrated in Section \ref{dataset-analysis}. 
More details can be found in Appendix A 1.1.
% MIR: Multi-Image Interleaved Reasoning dataset.
% The objective of the MIR is to enhance the ability of MLLMs to comprehend interleaved multi-image data.

\subsection{Dataset Overview}
\label{dataset-overview}
Focusing on interleaved multi-image reasoning, our MIR consists of $138,277$ images and $22,257$ multiple-choice questions. 
MIR enhances MLLMs' comprehension and performance in complex multimodal scenarios by analyzing intricate image-text relationships.
Existing multi-image benchmarks always focus on simple relationships between images (e.g., similarity, spatial location, or long-context reasoning). 
However, real-world reasoning tasks are far more complex, requiring models to recognize spatial relationships across multiple views or objects, infer sequential dependencies in dynamic temporal sequences or state changes, and perform logical reasoning based on counterintuitive or abstract concepts.
Thus we adopt \textit{Spatial}, \textit{Sequential}, and \textit{Analytical} classification framework as the foundation for data collection, and further refine it into $12$ specific subcategories according to the types of the collected data.

\begin{figure}[ht!]
    \centering
\includegraphics[width=1.0\linewidth]{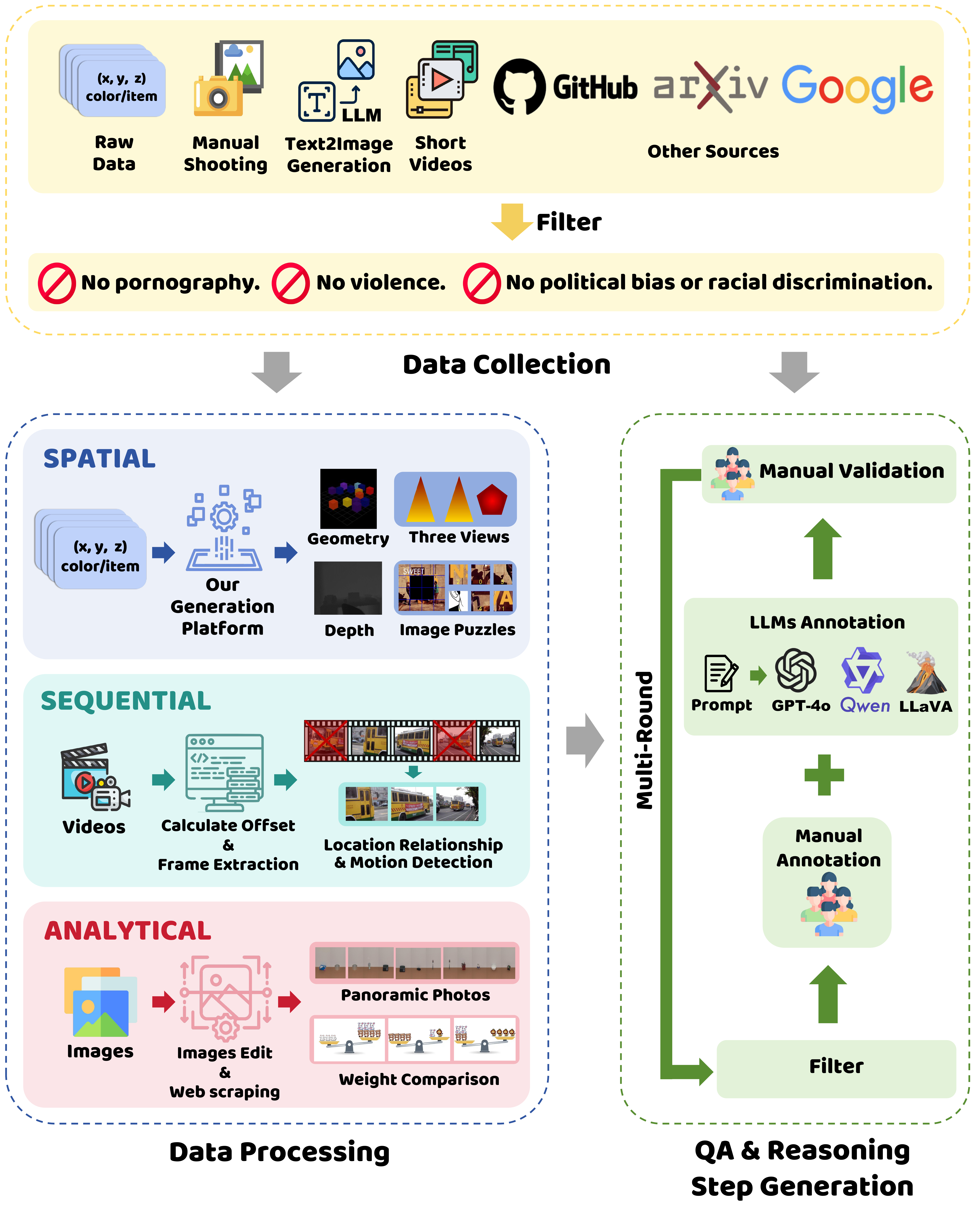}
    \caption{\textbf{Illustration of data construction.} We collect and filter data from multiple sources to obtain raw images or text. For spatial tasks, we generate/synthesize images by our platform; for sequential tasks, we extract frames/calculate offsets from videos; for analytical tasks, we leverage web scraping/model editing to process images. The final dataset is created by annotating images and text through multiple rounds using LLMs and human effort.}
    \label{fig:data_construction}
    \vspace{-5mm}
\end{figure}

\subsection{Dataset Constrution}
\label{dataset-construction}

% Although many Vision-Language Models (VLMs) have evolved to handle tasks involving the understanding of multiple images, they frequently encounter significant challenges, such as misalignment with instructions, errors in visual cognition, and inadequate skills in merging textual and visual data in multimodal contexts. These limitations hinder the deployment of multimodal models in real-world applications where seamless integration of text and images is crucial.

% By incorporating detailed reasoning chains and context comprehension processes, Our dataset aims to improve the model's ability to seamlessly integrate textual and visual information. This approach helps address instruction misalignment, minimizes visual cognition errors, and enhances multimodal understanding in complex interleaved contexts.

% In the meanwhile, existing multimodal datasets often focus on simple relationships between images (e.g., similarity, spatial location, or short-context reasoning) and provide concise answers that lack detailed reasoning steps. However, real-world multimodal reasoning tasks are far more complex, requiring models to:

% To address these challenges and gaps , we introduce MIR (Multimodal Interleaved Reasoning) , a novel dataset designed from a data-centric perspective . MIR is meticulously crafted to encapsulate the rich and intricate multimodal reasoning processes encountered in real-world scenarios, covering diverse formats and types of tasks. 
\textbf{Data Collection}: To mitigate risks of data leakage and ensure compliance, MIR adopts a multi-faceted data collection strategy. 
Specifically, the MIR comprises three primary sources: (1) Original recordings captured by our research team in controlled environments.
(2) Publicly accessible content curated from short-video platforms.
(3) Educational resources harvested from open-access learning platforms. 
All collected data undergoes a rigorous multi-stage review process, where sensitive materials including political content and explicit imagery are systematically filtered out. 
Furthermore, proper authorization and usage rights have been secured through official channels from both platform providers and content creators, ensuring full legal compliance and ethical data-handling practices.
More details of the collection process are available in the Appendix.

\noindent\textbf{Data Processing}:
As shown in Figure \ref{fig:data_construction}, we adopt a semi-automatic approach to generate Q\&A pairs within our benchmark.
Towards different task types, we employ distinct image acquisition and preprocessing strategies.
For Spatial tasks, we develop a three-dimensional geometric platform system. 
This platform is able to generate various geometric shapes (including cubes, cylinders, and cones) at any coordinate point in 3D space and automatically produce multi-view projections of these shapes, such as front, side, and top views.
For Sequential tasks, we mainly rely on video frame extraction and calculating the positional changes of objects across different images to generate data samples.
For Analytical tasks, we implement a dual-channel image acquisition strategy: (1) targeted web scraping from domain-specific sources to collect relevant real-world images, and (2) synthetic image generation using state-of-the-art text-to-image models \cite{qwen} to create tailored scene representations.

\noindent\textbf{Q\&A Generation}
Our approach primarily leverages LLMs for the automatic generation of question-answer pairs and reasoning steps. 
During the process of Q\&A Generation, we adopt different generation strategies tailored to the characteristics of various tasks: 
For tasks with a high degree of structure, we utilize predefined question templates, where specific entities or image region information could be directly embedded into the templates. 
For other tasks, we construct questions based on multimodal information sources, including image captions, prompts used for image generation, and contextual information related to the images. 
Throughout the question generation process, we place particular emphasis on ensuring logical consistency and multi-image coverage, guaranteeing that the generated questions thoroughly assess the model's ability to understand and reason across multiple images.
For structured questions, answers are derived through manual techniques, including coordinate calculations and logical reasoning. 
For unstructured questions, images are first converted into corresponding textual descriptions, which are then concatenated with the questions, and answers are generated using LLMs. 
Furthermore, to construct Multi-Choice Questions, we employ the LLM to generate three plausible yet incorrect answer options for each question, ensuring a comprehensive and challenging evaluation framework.

\noindent\textbf{Reasoning Step Generation}
We primarily generate reasoning steps by combining human expertise with LLMs. 
Specifically, we first annotate critical steps according to the task difficulty, such as captioning or text-to-image generation, and then use LLMs to assist in annotating the remaining steps. 
Finally, we manually inspect error-prone tasks (e.g.Forced Perspective, ScaleBlur).
This process is repeated multiple times to attain the final reasoning steps.
The requirements for each reasoning step are as follows: \textit{Summary}: MIR provides a high-level summary interpretation of the question, outlining the primary aspects of the problem it intends to address. \textit{Caption}: MIR delivers a brief and comprehensive summary of the images contained in each piece of data. \textit{Text to region}: MIR provides an in-depth examination of the methods for precisely aligning specific image regions with their associated textual descriptions. \textit{Region to region}: MIR further illustrates the logical relationships between various image regions. \textit{Conclusion}: MIR synthesizes an answer based on the preceding reasoning. 
Through this approach, we ultimately obtain MIR of high quality.
\vspace{-2mm}
\subsection{Dataset Analysis}
\label{dataset-analysis}
\begin{figure}[htbp]
    \centering
    % 第一张子图
    \begin{subfigure}[htbp]{0.45\textwidth}
        \centering
        \includegraphics[width=\textwidth]{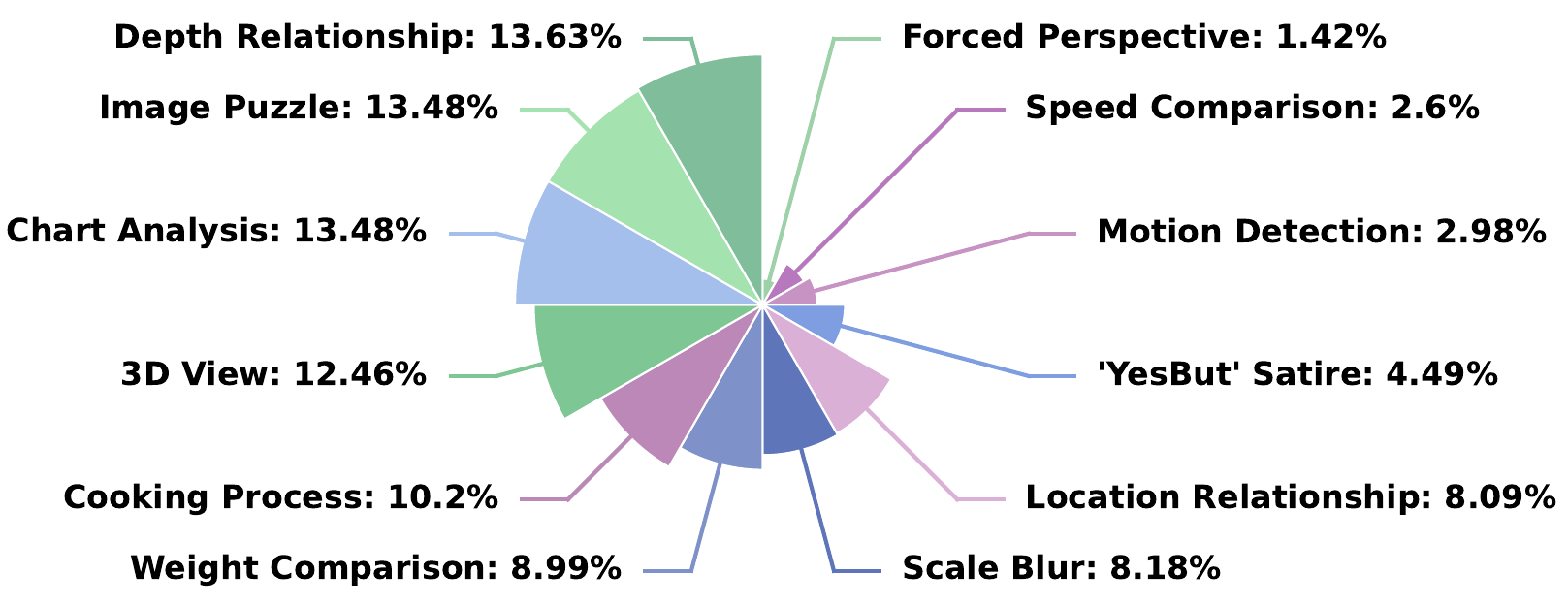}
        \caption{}
        \label{fig:pie_chart_1}
    \end{subfigure}
    \hfill  % 使子图之间有空隙
    % 第二张子图
    \begin{subfigure}[htbp]{0.45\textwidth}
        \centering
        \includegraphics[width=\textwidth]{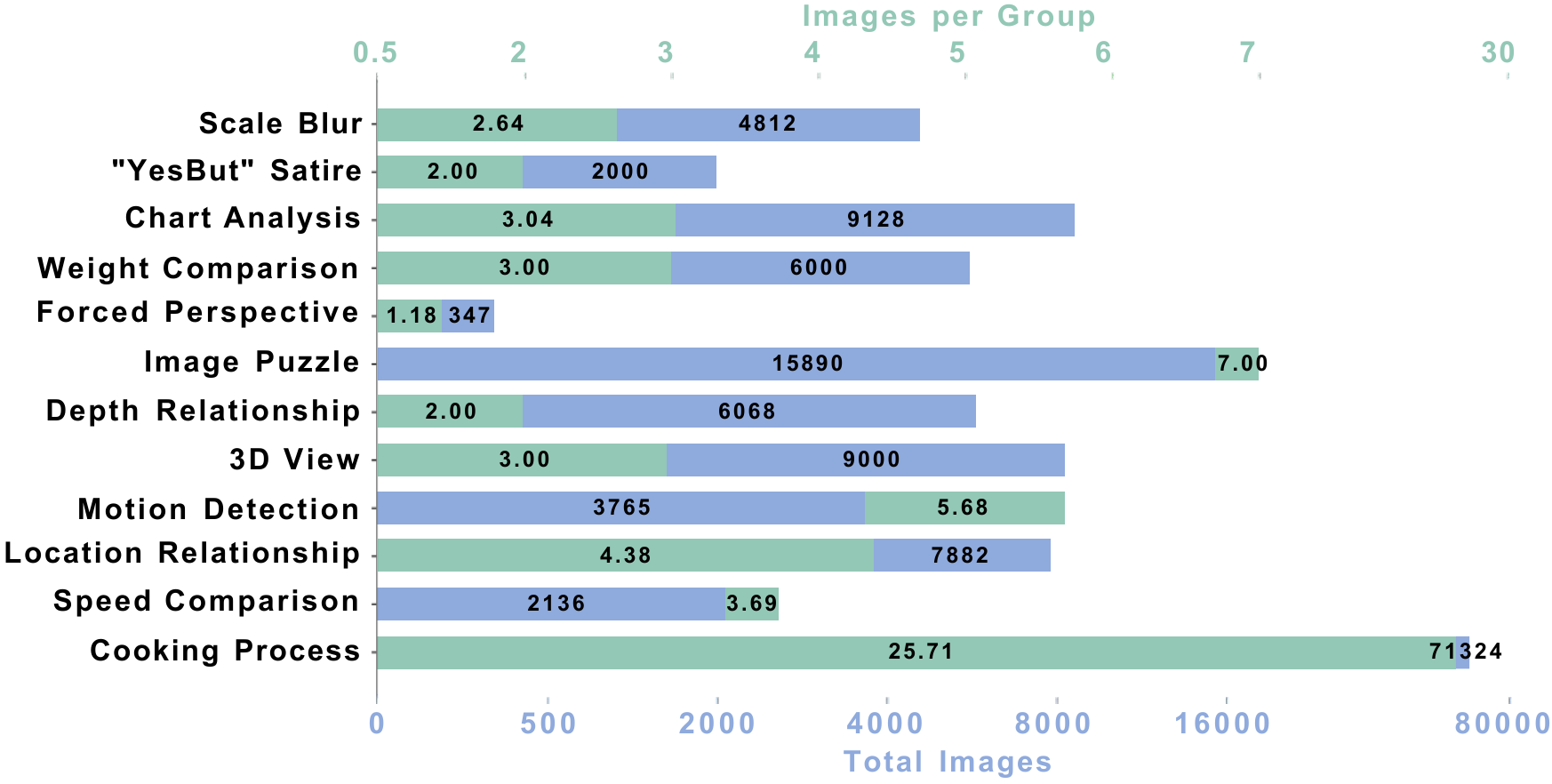}
        \caption{}
        \label{fig:statistics_of_image}
    \end{subfigure}
    % 第三张子图
    \begin{subfigure}[htbp]{0.45\textwidth}
        \centering
        \includegraphics[width=\textwidth]{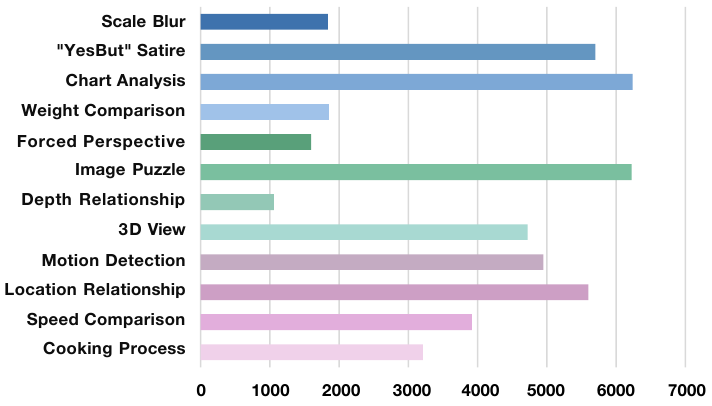}
        \caption{}
        \label{fig:statistics_of_chart}
    \end{subfigure}

    \caption{\textbf{Statistics of our MIR dataset.} Figure (a) shows the categories of MIR along with their respective proportions. Figure (b) illustrates the total number of images as well as the number of images per group for each task type. The average character count per annotation for each category is depicted in Figure (c)}
    \label{fig:multiple_subfigures}
    \vspace{-3mm}
\end{figure}

Our MIR is organized into three primary categories, each of which is further divided into four subcategories. The dataset consists of $138,377$ images and $22,257$ annotations, with each annotation averaging approximately $3,970$ characters. The distribution of annotation categories in the dataset is shown in Figure \ref{fig:pie_chart_1}, while the distribution of the number of images is presented in Figure \ref{fig:statistics_of_image}. Additionally, the average character count per annotation for each category is depicted in Figure \ref{fig:statistics_of_chart}. More details can be found in the Appendix.

\section{Methedology}
\label{Sec:Methedology}
\begin{figure*}[ht!]
    \centering
    \includegraphics[width=180mm]{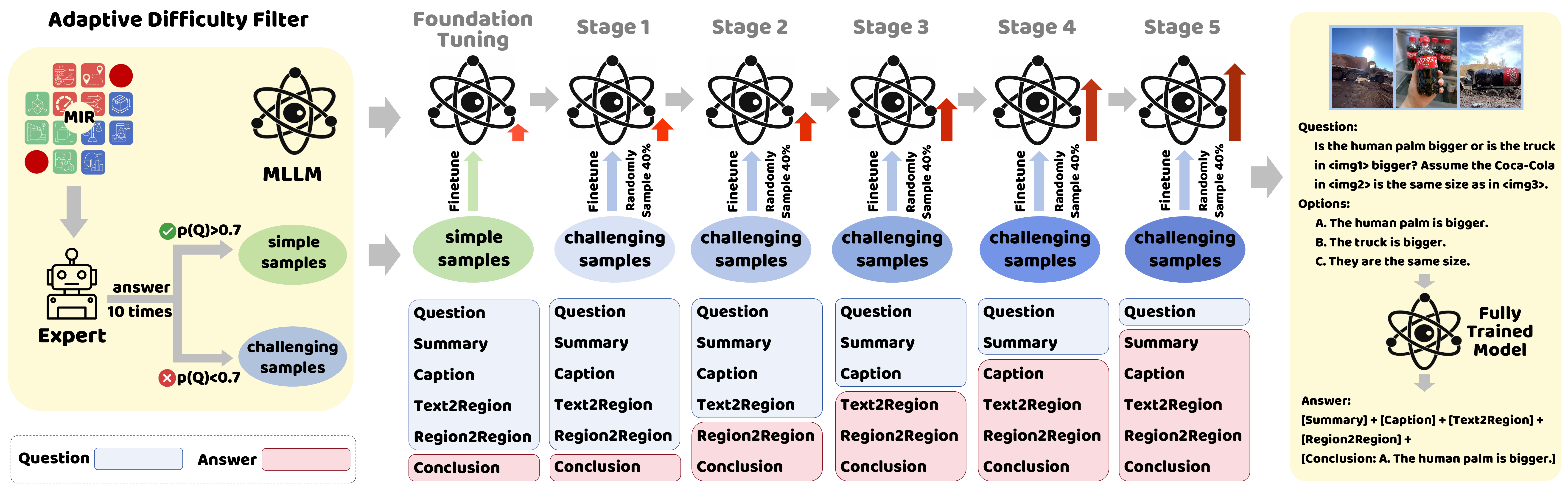}
    \caption{\textbf{Architecture of the proposed method.} We use an expert model to split the dataset into easy and difficult samples. After fine-tuning on easy samples, we sample 40\% of difficult ones for multi-stage curriculum learning. First, we combine the question with summary, caption, text2region, and region2region as input to generate the conclusion. Next, we move region2region to the target output, then add text-to-region from the question, followed by the caption, and finally the summary. The goal is for the model to autonomously generate a full reasoning process from the original question and arrive at the correct answer.}
    \label{fig:model}
    \vspace{-3mm}
\end{figure*}
In this section, we propose a novel method to fine-tune MLLMs based on curriculum learning, seamlessly integrating the reasoning steps of our MIR. 
As shown in Figure \ref{fig:model}, our method first employs a screening mechanism to classify all samples based on their difficulty. 
Then, we fine-tune the MLLMs with simple samples to establish a solid foundational understanding. 
Subsequently, more complex samples are employed to further refine the MLLMs, enhancing its capability to grasp deeper knowledge and solve intricate problems.
Unlike traditional fine-tuning methods that directly focus on question-answer pairs, our method aims to guide the model not only in providing correct answers but also in grasping the underlying thought process and problem-solving methodology. 
This approach enhances the MLLM's adaptability to tasks of varying complexities while simultaneously improving its generalization ability.
\subsection{Adaptive Difficulty Filter}
The adaptive difficulty filter aims to effectively distinguish between challenging and simple samples within MIR. 
Specifically, we input each question into the Qwen-VL2.5\cite{Qwen2.5-VL} model for inference and repeat this process $10$ times to obtain stable evaluation results. 
Based on the model's performance, we establish a clear classification criterion: if the correct response rate for a given question exceeds $70\%$, it is categorized as simple data; conversely, if the correct rate falls below $70\%$, it is classified as challenging data. 
This probability-based classification method ensures the objectivity and reliability of data categorization.
Here \( Q \) denotes a question within MIR, \( n = 10 \) is the number of tests for each question, \( c(Q) \) is the number of correct responses for question \( Q \) out of \( n \) trials, and \( p(Q) = \frac{c(Q)}{n} \) is the correct response rate for question \( Q \). The adaptive difficulty filter classifies \( Q \) as simple or challenging based on the following rule:
\begin{align}
    \text{Classification}(Q) = \begin{cases} \text{Simple}, & \text{if } p(Q) \geq 0.7 \\
    \text{Challenging}, & \text{if } p(Q) < 0.7 
    \end{cases}
\end{align}
\begin{table*}[!ht]
    \small
    \centering
    \begin{tabular}{lcccc|cccc}
    \toprule
        Method & \multicolumn{4}{c}{In-domain} & \multicolumn{3}{c}{Out-of-domain} \\ 
        \cmidrule(lr){2-5} \cmidrule(lr){6-8}
        ~ & Spatial & Sequential & Analytical & Average  & MIRBENCH & BLINK & MUIRBench \\ \midrule
        \addlinespace[1pt]
        Mantis & 29.30\% & 28.57\% & 33.12\% & 30.46\% & 42.83\% & 47.21\% & 37.53\% \\ 
        \addlinespace[1pt]
        Mantis(Tuning) & 33.21\% & 30.11\% & 36.78\% & 33.71\% & 43.10\% &  48.03\% & 38.58\% \\ 
        \addlinespace[1pt]
        Mantis(Ours) & \textbf{35.25\%} & \textbf{31.04\%} & \textbf{40.26\%} & \textbf{35.99\%} & \textbf{40.82\%} & \textbf{49.17\%} & \textbf{39.01\%}  \\ 
        \midrule
        LLava-Next-Interleave & 31.23\% & 22.74\% & 30.09\% & 28.79\% & 40.30\% & 48.87\% & 38.57\%  \\ 
        \addlinespace[1pt]
        LLava-Next-Interleave(Tuning) & 39.07\% & 25.01\% & 35.44\% & 34.42\% & 41.21\% & 49.31\% & \textbf{41.32\% }\\ 
        \addlinespace[1pt]
        LLava-Next-Interleave(Ours) & \textbf{42.25\%} & \textbf{25.04\%} & \textbf{37.85\%} & \textbf{36.58\%} & \textbf{43.19\%} & \textbf{52.27\%} & 40.25\% \\ 
        \midrule
        LLava-OneVision & 42.31\% & 36.53\% & 38.12\% & 39.31\% & 45.91\% & 49.00\% & 40.61\%  \\ 
        \addlinespace[1pt]
        LLava-OneVision(Tuning) & 45.17\% & 39.29\% & 41.20\% & 42.36\% & 47.04\% & 50.34\% & 42.22\%\\ 
        \addlinespace[1pt]
        LLava-OneVision(Ours) & \textbf{50.34\%} & \textbf{41.27\%} & \textbf{45.99\%} & \textbf{47.60\%} & \textbf{49.23\%} & \textbf{52.93\%} & \textbf{45.34\%} \\ 
        \midrule
        Qwen2-VL & 29.00\% & 56.60\% & 45.96\% & 40.44\%  & 50.04\% & 50.34\% & 45.67\% \\
        \addlinespace[1pt]
        Qwen2-VL(Tuning) & 37.31\% & 58.33\% & 49.59\%  & 45.15\% & 53.45\% & 52.19\% & 45.51\% \\
        \addlinespace[1pt]
         Qwen2-VL(Ours) & \textbf{40.42\%} & \textbf{59.65\%} & \textbf{52.93\%}  & \textbf{51.76\%} & \textbf{54.47\%} & \textbf{56.61\%} & \textbf{46.67\%} \\
         \midrule
        Mplug-Owl3 & 37.83\% & 28.65\% & 30.71\% & 33.13\% & 44.43\% & 49.90\% & 37.50\% \\ 
        \addlinespace[1pt]
        Mplug-Owl3(Tuning) & 40.34\% & 31.21\% & 35.66\% & 36.51\% & 46.09\% & 52.04\% & 37.93\%\\ 
        \addlinespace[1pt]
        Mplug-Owl3(Ours) & \textbf{44.17\%} & \textbf{34.90\%} & \textbf{39.64\%} & \textbf{40.36\%} & \textbf{49.31\%} & \textbf{55.57\%} & \textbf{39.25\%} \\ 
        \bottomrule
    \end{tabular}
    \caption{\textbf{Main results of our method.} We evaluate the performance of the base MLLMs, fine-tuned MLLMs, and MLLMs enhanced with our method across multiple benchmarks. The results show that nearly all MLLMs improved after fine-tuning with MIR, with our method delivering the most significant performance gains.}
    \label{main_result} 
    \vspace{-4mm}
\end{table*}
\subsection{Training Strategy}
Our training strategy is designed to progressively guide the model in mastering the ability to solve complex problems. 
To achieve this goal, we propose a progressive ``easy-to-hard'' training framework. 
Specifically, our framework consists of two parts: first, we fine-tune the model using simple samples to help it develop a foundational problem-solving capability.
Then, we adopt a stage-wise approach to gradually enhance the model’s reasoning ability.
At each training stage, we sample $40\%$ of the data from the pool of challenging examples and structurally refine them to align with the current training objectives. 
In the first stage, the input question \( Q \) is augmented by concatenating it with all reasoning steps \( S = \{s_1, s_2, \dots, s_n\} \), where \( n \) is the total number of steps. 
The answer \( A \) remains unchanged.
The new input \( Q_{\text{new}} \) and output \( A_{\text{new}} \) are defined as:
\begin{align}
Q_{\text{new}} = Q \oplus s_1 \oplus s_2 \oplus \dots \oplus s_n \quad\quad
A_{\text{new}} = A
\end{align}
Here, \( \oplus \) denotes the concatenation operation.
In the second stage, the input question \( Q \) is combined with all reasoning steps except the final step \( s_n \). The answer \( A \) aligns with the final reasoning step \( s_n \). The new question \( Q_{\text{new}} \) and the answer \( A_{\text{new}} \) are defined as:
\begin{align}
Q_{\text{new}} = Q \oplus s_1 \oplus s_2 \oplus \dots \oplus s_{n-1} \quad
A_{\text{new}} = A \oplus s_n 
\end{align}
In the later stages, we progressively reduce the guidance on reasoning steps. 
The final objective is for the model to autonomously generate a complete reasoning process based solely on the original question and reach the correct answer.
This design enables the model to transition from relying on external guidance to independently solving complex problems, achieving a fundamental shift from ``learning to reason'' to ``reasoning autonomously''.
\section{Experiments}
\label{Sec:Experiment}  
\begin{figure*}[ht!]
    \centering
    \includegraphics[width=1.0\linewidth]{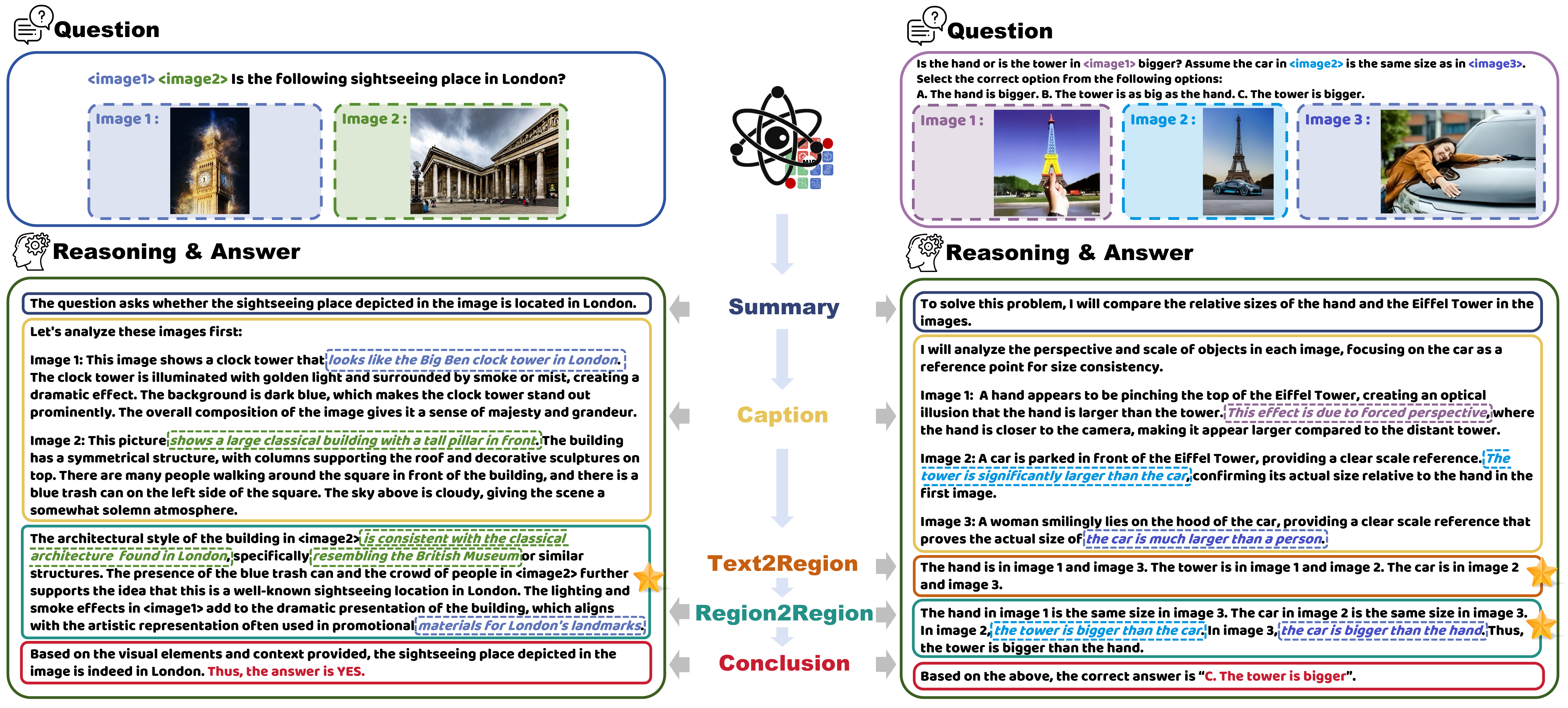}
    \caption{\textbf{Case Study of our method.} The model follows a structured reasoning process: it clarifies the question intent (summary), analyzes the image (caption), localizes the target region (text2region), establishes relationships between regions (region2region), and derives the correct answer (conclusion). These results demonstrate that the MLLM trained with our method is able to maintain a structured reasoning pipeline, enabling it to produce more accurate and logically consistent answers.}
    \label{fig:case_study}
    \vspace{-5mm}
\end{figure*}
\subsection{Implementation Details}
In the experimental setup section, we select current state-of-the-art open-source MLLMs for systematic evaluation.
All of these models possess functionality for multi-image interleaved comprehension.
Specifically, we choose five representative models for comparative experiments: Mantis\cite{jiang2024mantis}, mPLUG-Owl3\cite{ye2024mplugowl3}, LLaVA-NeXT-Interleave\cite{llava-next-interleave}, Qwen2-VL\cite{Qwen2-VL} and LLaVA-OneVision\cite{LLaVA-OneVision}. 
For each model, we implement three training strategies: (1) Zero-shot evaluation using the original pre-trained model, (2) Full Fine-tuning, (3) Fine-tuning by our proposed method.
To ensure the generalizability of our method. the evaluation covers both in-domain and out-of-domain perspectives.
For in-domain evaluation, we assess model performance on our custom dataset in three dimensions: sequence analysis, spatial relations, and analytical inference. 
For out-domain evaluation, we employ three cross-domain benchmarks, including MIRB, BLINK, and MUIR. 
During the evaluation phase, since lmms-eval\cite{lmms_eval2024} relies on rule-based answer matching, long outputs often struggle to align with the correct answers. 
Therefore, we uniformly employ GPT to match model outputs with the ground truth.
This evaluation scheme ensures the reliability and comprehensiveness of the experimental results.
\subsection{Results of Our Method}
The experiment results are illustrated in Table \ref{main_result}.
We conduct comprehensive evaluations of the original MLLMs, the fine-tuned MLLMs, and the MLLMs enhanced with our proposed method across multiple benchmarks. 
During the tests of in-domain, the models demonstrate a notable improvement after finetuning in in-domain testing, with an increase of approximately $3\%$ to $5\%$. 
By employing our method, the model's performance was further enhanced, achieving a rise of $7\%$.
During the tests of out-of-domain, by adopting our proposed method, most models achieve a performance improvement of $1\% $to $5\%$ over the baseline. 
In contrast, traditional fine-tuning methods yields a maximum gain of only $2\%$. 
The experimental results demonstrate that MLLMs fine-tuned with our method achieve significant performance improvements across all benchmarks, fully validating the effectiveness of MIR. 
Notably, compared to using MIR alone, our approach lead to even more substantial performance gains. 
This outcome strongly confirms that our designed ``easy-to-hard‘’ learning strategy effectively guides the model to progressively master complex features, thereby achieving better generalization capabilities. \\
\indent Through experiments, we have discovered some interesting phenomena: (1) The ability to follow instructions is critical for step-wise curriculum learning, as stronger instruction adherence enhances the effectiveness of the learning process. 
(2) The model shows the least improvement in performance on sequence tasks after fine-tuning, which may be attributed to the fact that the images in sequential tasks are extracted from videos. 
These extracted images often fail to accurately represent the entire video, leading to the model's suboptimal performance in sequential tasks.
\begin{table}[!ht]
    \centering
    \resizebox{1.0\linewidth}{!}{
    \begin{tabular}{l*{5}{c}} % 使用 *{5}{c} 来简化列格式定义
        \toprule
        Qwen2-VL & \textbf{stage1} & \textbf{stage2} & \textbf{stage3} & \textbf{stage4} & \textbf{stage5} \\ 
        \midrule
        spatial & 35.19\% & 34.60\% & 40.57\%  & 46.29\% & 50.96\% \\ 
        \addlinespace[1pt]
        Sequential & 32.22\% & 35.55\% & 33.05\% & 34.44\% & 36.66\% \\ 
        \addlinespace[1pt]
       Analytical & 44.85\% & 38.97\% & 43.52\%  & 48.52\% &  58.82\% \\ 
       \addlinespace[1pt]
       Average & 38.01\% & 36.73\% & 41.03\%  & 45.97\% & 51.76\%\\ 
        \bottomrule
    \end{tabular}
    }
    \caption{\textbf{Ablation study.} This table presents the Qwen2-VL's performance on MIR after completing training at each stage.}
    \label{tab:ablation_study}
    \vspace{-4mm}
\end{table}

\subsection{Ablation Study}
As illustrated in the Table \ref{tab:ablation_study}.
The decline in Stage 2 stems from higher task complexity, requiring more training data or finer adjustments to adapt to the logically intensive structure. 
In contrast, Stage 3 and Stage 4 tasks align better with the model's reasoning capabilities, leading to performance recovery. Stage 4, in particular, closely aligns with the model's core strengths, driving significant improvement.

\subsection{Case Study}
The first case is the output of mPLUG-Owl3 on the sightseeing task in MIRB\cite{wang2024muirbench} after tuning with our method. 
As shown in Figure \ref{fig:case_study}, mPLUG-Owl3 demonstrates a complete reasoning process: it first clarifies the intent of the question (summary), then conducts a detailed analysis of the image content (caption), effectively establishes relationships between images (region2region), and ultimately derives the correct answer (conclusion).
Since this data is not strictly interleaved, the text-to-region process is absent.
The second case aims to compare the sizes of a tower and a car. 
The model first formulates its reasoning process (summary). 
It then thoroughly analyzes each image (caption), identifies and localizes the objects to be compared (text2region), and establishes their relationships (region2region). 
Finally, leveraging these relationships, the model derives the correct answer (conclusion).
These cases illustrate that the MLLM trained with our method is able to follow a structured reasoning process and ultimately arrive at the correct answer.
\section{Conclusion}
\label{Sec:conlusion}
This paper introduces MIR, a novel benchmark for multi-image interleaved reasoning. 
Each data instance includes a detailed reasoning step, designed to guide models in understanding the intricate relationships between multiple images and interleaved texts.
Building on these reasoning steps, we propose a stage-wise curriculum learning approach to enhance models' reasoning and generalization abilities through progressive learning.
We believe MIR will inspire further exploration and development in various downstream tasks, such as cross-modal reasoning, scene understanding, and complex visual-textual analysis. 
\section*{Acknowledgement} \label{sec:Acknowledgement}
This work was supported in part by the National Key Research and Development Program of China under Grant 2022YFB2902200.
This work was supported in part by the National Natural Science Foundation of China under Grant 62471064.
{
    \small
    \bibliographystyle{ieeenat_fullname}
    \bibliography{main}
}

\end{document}